\documentclass[sigconf]{acmart}
\usepackage{float}
\usepackage{multirow}
\usepackage{CJKutf8}
\usepackage{balance}
\usepackage{graphicx}
\graphicspath{{}} 
\AtBeginDocument{%
  \providecommand\BibTeX{{%
    \normalfont B\kern-0.5em{\scshape i\kern-0.25em b}\kern-0.8em\TeX}}}

\copyrightyear{2021}
\acmYear{2021}
\setcopyright{rightsretained}
\acmConference[FAccT '21]{Conference on Fairness, Accountability, and Transparency}{March 3--10, 2021}{Virtual Event, Canada}
\acmBooktitle{Conference on Fairness, Accountability, and Transparency (FAccT '21), March 3--10, 2021, Virtual Event, Canada}\acmDOI{10.1145/3442188.3445916}
\acmISBN{978-1-4503-8309-7/21/03}
\begin{document}

\title[Censorship of Online Encyclopedias]{Censorship of Online Encyclopedias: Implications for NLP Models}

%%\orcid{1234-5678-9012}
\author{Eddie Yang}
\authornote{Both authors contributed equally to this research.}
\email{z5yang@ucsd.edu}
\affiliation{%
  \institution{University of California, San Diego}
  \streetaddress{9500 Gilman Dr}
  \city{La Jolla}
  \state{California}
  \postcode{92093}
}

\author{Margaret E. Roberts}
\authornotemark[1]
\email{meroberts@ucsd.edu}
\affiliation{%
  \institution{University of California, San Diego}
  \streetaddress{9500 Gilman Dr}
  \city{La Jolla}
  \state{California}
  \postcode{92093}
}

\renewcommand{\shortauthors}{Yang and Roberts}

\begin{abstract}
  While artificial intelligence provides the backbone for many tools people use around the world, recent work has brought to attention that the algorithms powering AI are not free of politics, stereotypes, and bias. While most work in this area has focused on the ways in which AI can exacerbate existing inequalities and discrimination, very little work has studied how governments actively shape training data. We describe how censorship has affected the development of Wikipedia corpuses, text data which are regularly used for pre-trained inputs into NLP algorithms. We show that word embeddings trained on Baidu Baike, an online Chinese encyclopedia, have very different associations between adjectives and a range of concepts about democracy, freedom, collective action, equality, and people and historical events in China than its regularly blocked but uncensored counterpart -- Chinese language Wikipedia. We examine the implications of these discrepancies by studying their use in downstream AI applications. Our paper shows how government repression, censorship, and self-censorship may impact training data and the applications that draw from them.
\end{abstract}



%%  \includegraphics[width=\textwidth]{sampleteaser}
%%  \caption{Seattle Mariners at Spring Training, 2010.}
%%  \Description{Enjoying the baseball game from the third-base
%%  seats. Ichiro Suzuki preparing to bat.}

\maketitle

\section{Introduction}
Natural language processing (NLP) as a branch of artificial intelligence provides the basis for many tools people around the world use daily. NLP impacts how firms provide products to users, content individuals receive through search and social media, and how individuals interact with news and emails. Despite the growing importance of NLP algorithms in shaping our lives, recently scholars, policymakers, and the business community have raised the alarm of how gender and racial biases may be baked into these algorithms. Because they are trained on human data, the algorithms themselves can replicate implicit and explicit human biases and aggravate discrimination \citep{sweeney2013discrimination,bolukbasi2016man, caliskan2017semantics}. Additionally, training data that over-represents a subset of the population may do a worse job at predicting outcomes for other groups in the population \citep{dressel2018accuracy}. When these algorithms are used in real world applications, they can perpetuate inequalities and cause real harm.

While most of the work in this area has focused on bias and discrimination, we bring to light another way in which NLP may be affected by the institutions that impact the data that they feed off of. We describe how censorship has affected the development of online encyclopedia corpuses that are often used as training data for NLP algorithms. The Chinese government has regularly blocked Chinese language Wikipedia from operating in China, and mainland Chinese Internet users are more likely to use an alternative Wikipedia-like website, Baidu Baike. The institution of censorship has weakened Chinese language Wikipedia, which is now several times smaller than Baidu Baike, and made Baidu Baike - which is subject to pre-censorship - an attractive source of training data. Using methods from the literature on gender discrimination in word embeddings, we show that Chinese word embeddings trained with the same method but separately on these two corpuses reflect the political censorship of these corpuses, treating the concepts of democracy, freedom, collective action, equality, people and historical events in China significantly differently.

%After establishing that these two corpuses reflect different word associations, we then use survey data to compare how such differences are viewed among mainland Chinese Internet users. We design a survey experiment where we present respondents in mainland China with a target word (for example ``democracy'') and the adjective closest to that target word from both Wikipedia and Baidu embeddings. We ask them which adjective better describes the target word. We then compare whether Wikipedia and Baidu word embeddings better reflect the word associations of individuals in mainland China. We find that adjectives from neither corpus are consistently preferred, indicating that the associations in the Baidu Baike corpus may not simply be a reflection of the views of mainland Chinese Internet users.

After establishing that these two corpuses reflect different word associations, we demonstrate the potential real-world impact of training data politics by using the two sets of word embeddings in a transfer learning task to classify the sentiment of news headlines. We find that models trained on the same data but using different pre-trained word embeddings make significantly different predictions of the valence of headlines containing words pertaining to freedom, democracy, elections, collective action, social control, political figures, the CCP, and historical events. These results suggest that censorship could have downstream effects on AI applications, which merit future research and investigation.

Our paper proceeds as follows. We first describe the background of how Wikipedia corpuses came to be used as training data for word embeddings and how censorship impacts these corpuses. Second, we describe our results of how word associations from Wikipedia and Baidu Baike word embeddings differ on concepts that pertain to democracy, equality, freedom, collective action and historical people and events in China. Last, we show that these embeddings have downstream implications for AI models using a sentiment prediction task.

\section{Pre-Trained Word Embeddings and Wikipedia Corpuses}

NLP algorithms rely on numerical representations of text as a basis for modeling the relationship between that text and an outcome. Many NLP algorithms use ``word embeddings" to represent text, where each word in a corpus is represented as a k-dimensional vector that encodes the relationship between that word and other words through the distance between them in k-dimensional space. Words that frequently co-occur are closer in space. Popular algorithms such as Glove \citep{pennington2014glove} and Word2Vec \citep{mikolov2013distributed} are used to estimate embeddings for any given corpus of text. The word embeddings are then used as numerical representations of input texts, which are then related through a statistical classifier to an outcome.

In comparison to other numerical representations of text, word embeddings are useful because they communicate the relationships between words. The bag-of-words representation of text, which represents each word as simply being included or not included in the text, does not encode the relationship between words -- each word is equidistant from the other. Word embeddings, on the other hand, communicates to the model which words tend to co-occur, thus providing the model with information that words like ``purse" and ``handbag'' as more likely substitutes than ``purse" and ``airplane".

Word embeddings are also useful because they can be pre-trained on large corpuses of text like Wikipedia or Common Crawl, and these pre-trained embeddings can then be used as an initial layer in applications that may have less training data. Pre-trained word embeddings have been shown to achieve higher accuracy faster \citep{qi2018and}. While training on large corpuses is expensive, companies and research groups have made available pre-trained word embeddings -- typically on large corpuses like Wikipedia or Common Crawl -- that can then be downloaded and used in any application in that language.\footnote{For example, Facebook's provides word embeddings in 294 languages trained on Wikipedia (\url{https://fasttext.cc/docs/en/pretrained-vectors.html} \citep{bojanowski2017enriching}.}

The motivation behind using pre-trained word embeddings is that they can reflect how words are commonly used in a particular language. Indeed, \citet{spirling2019word} show that pre-trained word embeddings do surprisingly well on a ``Turing test" where human coders often cannot distinguish between close words produced by the embeddings and those produced by other humans. To this end, Wikipedia corpuses are commonly selected to train word embeddings because they are user-generated, open-source, cover a wide range of topics, and are very large.\footnote{A Google Scholar search of ``pre-trained word embeddings" and Wikipedia returns over 2,000 search results as of January 2021.}

At the same time as pre-trained embeddings have become popular for computer scientists in achieving better performance for NLP tasks, some scholars have pointed to potential harms these embeddings could create by encoding existing biases into the representation. The primary concern is that embeddings replicate existing human biases and stereotypes in language and using them in downstream applications can perpetuate these biases (see \citet{Sun_2019} for a review). \citet{caliskan2017semantics} show that word embeddings reflect human biases, in that associations of words in trained word embeddings mirror implicit association tests. Using simple analogies within word embeddings, \citet{bolukbasi2016man}, \citet{garg2018word}, and \citet{manzini2019black} show that word embeddings can encode racial and gender stereotypes. While these word associations can be of interest to social science researchers, they may cause harm if used in downstream tasks \citep{barocas2016big, papakyriakopoulos2020bias}.

More generally, research in machine learning has been criticized for not paying enough attention to the origin of training datasets and the social processes that generate them \citep{geiger2020garbage}. Imbalances in the content of training data have been shown to create differential error rates across groups in areas ranging from computer vision to speech recognition \citep{tatman2017gender, torralba2011unbiased}. Some scholars have argued that training datasets should be representative of the population that the algorithm is applied to \citep{shankar2017no}. 

\section{Censorship of Chinese Language Wikipedia and Implications for Chinese Langauge NLP} 

We consider another mechanism through which institutional and societal forces impact the corpuses that are used to train word embeddings: government censorship. While we use the example of online encyclopedias and word embeddings to make our point, its implications are much more general. Government censorship of social media, news, and websites directly affects large corpuses of text by blocking users' access, deleting individual messages, adding content through propaganda, or inducing self-censorship through intimidation and laws \citep{deibert2008access,morozov2012net, MacKinnon12, king2013censorship, king2017chinese,sanovich2018turning, Roberts18}. 

While Wikipedia's global reach makes it an attractive corpus for training models in many different languages, Wikipedia has also been periodically censored by many governments, including Iran, China, Uzbekistan, and Turkey \citep{clark2017analyzing}.  China has had the most extensive and long-lasting censorship of Wikipedia.  Chinese language Wikipedia has been blocked intermittently ever since it was first established in 2001.  Since May 19, 2015, all of Chinese language Wikipedia has been blocked by the Great Firewall of China \citep{Welinder2015,Oberhaus2015}.  More recently, not just Chinese language Wikipedia, but all language versions of Wikipedia have been blocked from mainland China \citep{Wikipedia2019}.

Censorship has weakened Chinese language Wikipedia by decreasing the size of its audience.  \citet{panandroberts19} estimate that the block of Chinese language Wikipedia in 2015 decreased page views of the website by around 3 million views per day. \citet{zhang2011group} use the 2005 block of Wikipedia to show that the block decreased views of Chinese language Wikipedia, which in turn decreased user contributions to Wikipedia not only from blocked users in mainland China, but also from unblocked users what had fewer incentives to contribute after the block.  While mainland Chinese Internet users can access Chinese language Wikipedia with a Virtual Private Network (VPN), evidence suggests that very few do \citep{chenyang2019, Roberts18}.  

Censorship of Chinese language Wikipedia has strengthened its unblocked substitute, Baidu Baike.  A similar Wikipedia-like website, Baidu Baike as of 2019 boasted 16 million entries, 16 times larger than Chinese language Wikipedia \citep{Zhang2019}. Yet, as with all companies operating in China, Baidu Baike is subject to internal censorship that impacts whether and how certain entries are written.  While edits to Chinese language Wikipedia pages are posted immediately, any edits to Baidu Baike pages go through pre-publication review.  While editors of Wikipedia can be anonymous, editors of Baidu Baike must register their real names. Additional scrutiny is given to sensitive pages, such as national leaders, political figures, political information, and the military, where Baidu Baike regulations stipulate that only government media outlets such as \emph{Xinhua} and \emph{People's Daily} can be used as sources.\footnote{See instructions at: \url{https://baike.baidu.com/item/\%E7\%99\%BE\%E5\%BA\%A6\%E7\%99\%BE\%E7\%A7\%91\%EF\%BC\%9A\%E5\%8F\%82\%E8\%80\%83\%E8\%B5\%84\%E6\%96\%99}.}

Pre-censorship of Baidu Baike affects the types of pages available on Baidu Baike and the way these pages are written.  While it's impossible to know without an internal list the extent to which missing pages in Baidu Baike are a direct result of government censorship, a substantial list of historical events covered on Chinese language Wikipedia including ``June 4th Incident'' and ``Democracy Wall'' and well-known activists such as Chen Guangcheng and Wu'erkaixi have no Baidu Baike page \citep{Ng2013}. For example, when we attempted to create entries on Baidu Baike such as ``June Fourth Movement" or ``Wu'erkaixi," we were automatically returned an error. 

Perhaps because of the size difference between the two corpuses, increasingly researchers developing cutting edge Chinese langauge NLP models are drawing on the Baidu Baike corpus \citep{sun2019ernie, wei2019nezha}.  Baidu Baike word embeddings have been shown to perform better on certain tasks \citep{li2018analogical}. Here, we assess the downstream implications of this choice on the representation of democratic concepts, social control, and historical events and figures.  First, we follow \citet{caliskan2017semantics} to compare the distance between these concepts and a list of adjectives and sentiment words.  Then, we show the downstream consequences of the choice of corpus on a predictive task of the sentiment of headlines.

\section{Distance from Democracy: Comparison Between Baidu Baike and Wikipedia Embeddings\label{sec:distance}}

In this section, we consider the differences in word associations among word embeddings trained with Chinese language Wikipedia and Baidu Baike.  We use word embeddings made available by \citet{li2018analogical}.\footnote{\url{https://github.com/Embedding/Chinese-Word-Vectors}} \citet{li2018analogical} train 300-dimensional word embeddings on both Baidu Baike and Chinese language Wikipedia using the same algorithm, Word2Vec \citep{mikolov2013distributed}.  For a benchmark, we also compare these two sets of embeddings to embeddings trained on articles from the \emph{People's Daily} from 1947-2016, the Chinese government's mouthpiece.\footnote{Also trained by \citet{li2018analogical} and made available at \url{https://github.com/Embedding/Chinese-Word-Vectors}.}  %These embeddings have become tremendously popular Chinese word representations, yielding $7,500$ ``stars" and $1,800$ ``forks" on GitHub.\footnote{\citet{li2018analogical}'s paper has received more than a hundred citations since its publication in 2018.}

To evaluate word associations, we follow \citet{caliskan2017semantics} and \citet{rodman2019timely} to compare the distance between a set of target words and attribute words to establish their relationships in each embedding space. Figure 1 gives a simplified graphical representation of the evaluation procedure in a 2-dimensional space. In this simple example, we might be interested in the position of a target word -- a concept we are interested in -- relative to a positive attribute word and a negative attribute word. For example, we can evaluate whether democratic concepts are represented more positively or negatively by comparing the angle between the vector for the target word ``Democracy" (in black) and a positive attribute word ``Stability" as well as a negative attribute word ``Chaos" (both in blue).

\begin{figure}[htbp]
	\centering
	\includegraphics[width=0.9\linewidth]{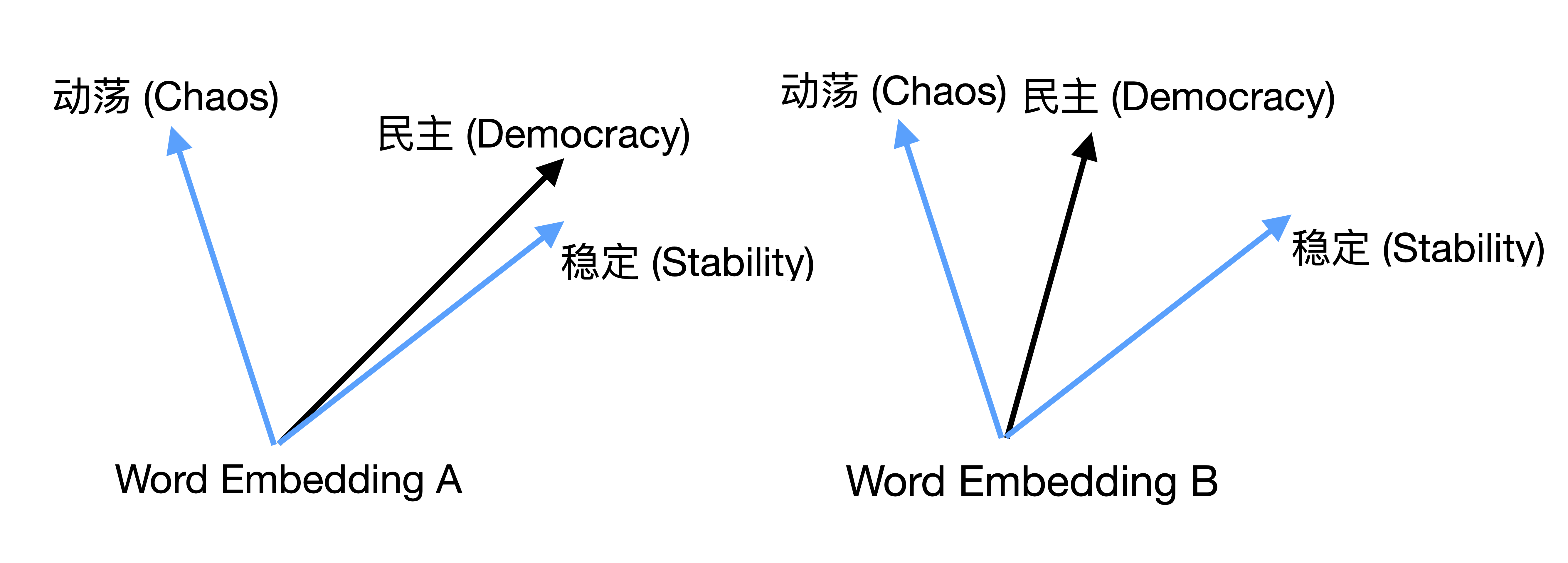}
	\Description{2-Dimensional word embeddings with ``democracy" as target word and ``chaose" and ``stability" as attribute words}
	\caption{Example of Word Embedding Comparison}
	\label{fig:embedding}
\end{figure}

In Figure \ref{fig:embedding}, ``Democracy'' in word embedding A has a more positive connotation than in word embedding B, because the relative position of the word ``Democracy" in embedding A with respect to the positive attribute word ``Stability" and the negative attribute word ``Chaos" is closer to the positive attribute word than ``Democracy" is in embedding B. To minimize the particularities of a single word and hence the variability of the result, we repeat this evaluation procedure across multiple target words representing the same concept (e.g. democracy) and compare them with multiple attribute words. In the next sections, we explain how we select target words, attribute words, how we pre-process the embedding space, and our results.

\subsection{Identifying Target Words}

We begin by delineating the categories of interest. In general, there are two broad categories we are interested in: 1) democratic political concepts and ideas and 2) known targets of propaganda. Based on past work, we know entries that fall under these categories have been the target of content control on Baidu Baike \citep{Ng2013}.Additionally, the first category captures ideas that we think are normatively desirable but discouraged in China. The second category captures the extent that the embeddings are consistent with propaganda.

For the first category, we include
\begin{enumerate}
\item Democratic values, in particular freedom and equality of rights.
\item Procedures of democracy, in particular features pertaining to elections.
\item Channels for voicing preferences in the form of collective actions such as protests and petitions.
\end{enumerate}

\noindent For the second category, we include

\begin{enumerate}
\item Social control, especially concepts related to repression and surveillance.
\item The Chinese Communist Party (CCP) and related features.
\item Significant historical events in China that involved the CCP, such as the Cultural Revolution.
\item Important figures who are extolled by the CCP.
\item Figures who are denounced by the CCP, such as political dissenters.
\end{enumerate}

For each of these categories, we do not want to select only one target word of interest, but rather a group of related words that all cover the same concept. We select a group of target words that “represent” this category as follows:

\begin{enumerate}
\item For categories other than historical events and negative figures, we first select a Chinese word that most closely represents the category of interest.\footnote{We asked three Chinese speakers to independently come up with the representative words and had them agree on a single word for each category. This step was done before analysis was performed.} For example, for the category of procedures of democracy, the Chinese word ``election" is selected.
\item We then calculate the cosine similarity of the representative word with all other words from the word embedding spaces (Wikipedia \& Baidu Baike).
\item From each corpus, we select 50 words that are closest to the representative word (words with the highest cosine similarity).
\item Of the 100 words closest to the representative word for each category, we include all words that could be thought to be synonymous or a subset of the more general category.  We drop those that are domain specific; for example, of the words for the category of procedures of democracy, we dropped the word "Japanese Diet", which is specific to the Japanese political system.
\item For categories on historical events and negative figures, we simply used the name of the person or of the historical event.
\item The full list of words for each category is presented in Appendix D.
\end{enumerate}

We opt for the data-driven approach in (3) and (4) to select target words in order to limit researcher degree of freedom. Furthermore, the selection of representative words in (1) and the pruning of synonyms in (4) were done by three native Chinese speakers to ensure the selected words provide good coverage of how the categories of interest are discussed in the Chinese context. 
%(Note: Currently, the above procedure is only done with the Wikipedia and Baidu Baike embedding spaces but not with the People’s Daily embedding space.)

\subsection{Selecting Attribute Words}

We use two strategies for selecting attribute words.  First, we draw on the literature on propaganda in China to select a set of positive and negative words that would be consistent with what we know about CCP propaganda narratives.  As scholars of propaganda have pointed out, the CCP has actively tried to promote the image of itself and China's political system as stable and prosperous, while characterizing Western democratic systems as chaotic and in economic decline \citep{brady2015authoritarianism, zhang2019china,Economist2016}.  Therefore, for our first set of words, which we call ``Propaganda Attributes Words," positive words include synonyms of stability and prosperity, while negative attribute words include synonyms of chaos, decline, and instability.  The full list for the set of propaganda attribute words is presented in Appendix E.

%We come up with two sets of attribute words to triangulate how each of the target word identified in the previous section is portrayed in each corpus. For the first set, we start with adjective words that mean either "stable" or "prosperous". We then include antonyms of these words. The idea is that, to bolster the performance legitimacy of the regime, targets of propaganda such as the CCP and social control should be intentionally associated more with stability and prosperity. On the other hand, concepts that run counter to the ideology of the CCP should be intentionally associated more with instability and poverty. The full list for the set of adjective words is presented in Appendix B. For subsequent discussions, we refer to this list of attribute words the adjective attributes.

For the second set of words, we are interested in whether the target words are more generally evaluated differently between the two corpuses. To test this, we make use of a dictionary of evaluative words specifically designed for Chinese natural language processing \citep{wang2016antusd}. The dictionary codes whether an evaluative word is positive, negative, or neutral. We follow the preprocessing instructions by \citet{wang2016antusd} by dropping all neutral words and only using the list of positive and negative evaluative words. A sample of the set of evaluative words is presented in Appendix F. For subsequent discussions, we refer to this list of attribute words as the ``Evaluative Attribute Words."

\subsection{Pre-processing Word Embedding Spaces}

There are two notable challenges when comparing different word embeddings. One, word embeddings produced by stochastic algorithms such as Word2Vec will embed words in non-aligned spaces defined by different basis vectors. This precludes naive comparison of word distances across distinct corpuses \citep{hamilton2016diachronic, rodman2019timely}. If the centroids of the two word embeddings are different, then using cosine similarity (i.e. the cosine of the angle between two vectors) to compare word associations across different corpuses can yield uninterpretable result. Figure 2 presents a simplified example of this problem. One word embedding, by virtue of being further away from the origin, yields a smaller angle between the two vectors, even though the relative positions of the two vectors in the two word embeddings are the same. 

To solve this problem, we standardize the basis vectors of each word embeddings by subtracting the means and dividing by the standard deviations of the basis vectors, so that each word embedding is centered around the origin with dimension length 1.
\begin{figure}[htbp]
	\centering
	\includegraphics[width=0.8\linewidth]{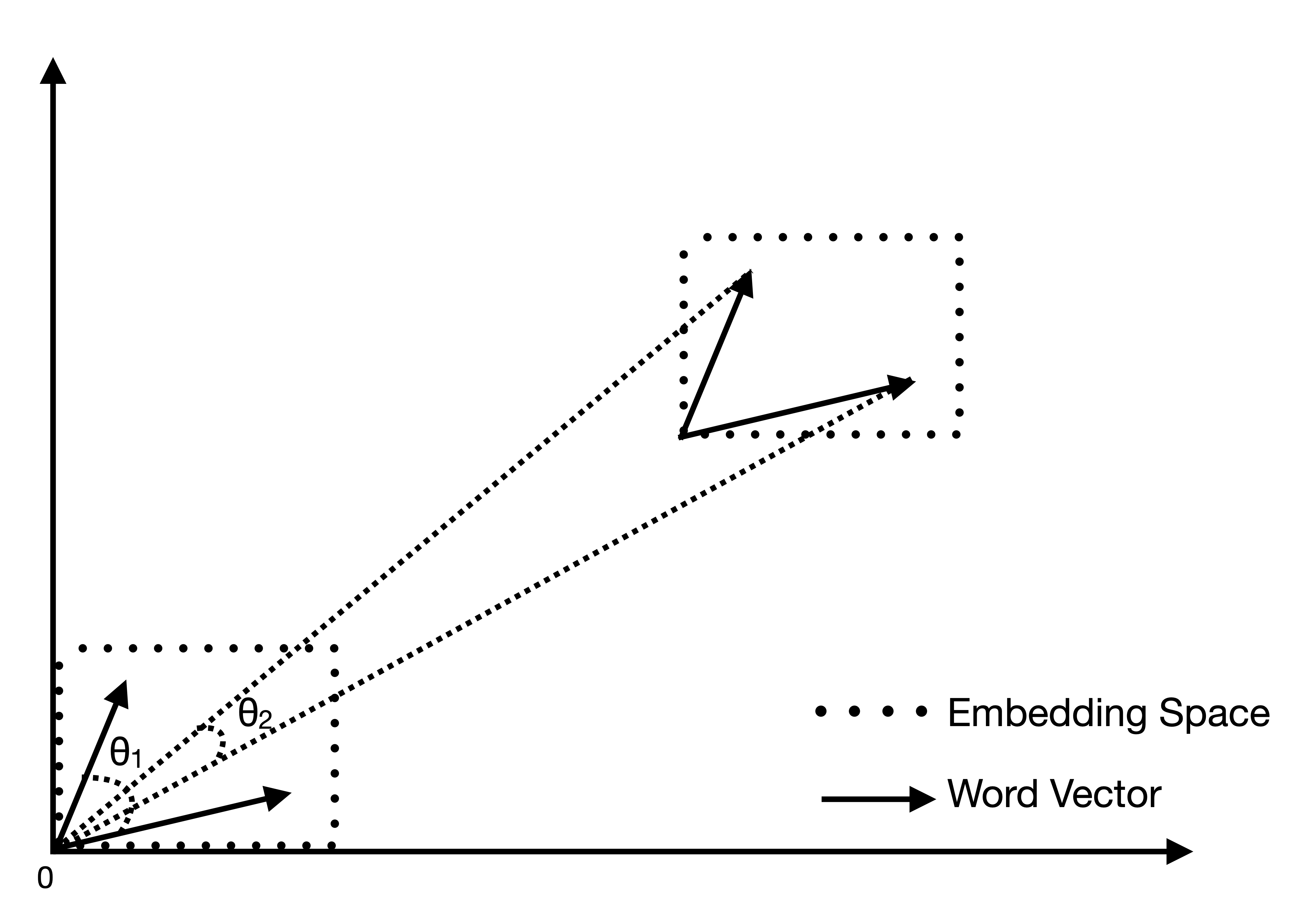}
	\Description{An example of nonalignment between two word embeddings}
	\caption{Nonalignment between Two Word Embeddings}
	\label{fig:alignment}
\end{figure}

Another problem is that word embeddings trained on different corpuses can have different vocabulary. This precludes us from comparing words that appear in one word embedding but are not present in the other word embedding. Because of this, we only keep the intersection of the vocabularies of word embeddings. As a result, six target words were dropped in the comparison between Wikipedia- and Baidu Baike-trained word embeddings and five target words were dropped in the comparison between Wikipedia- and \emph{People's Daily}-trained word embeddings.

\subsection{Expectations}

We expect ideas that are normatively appealing but discouraged in China to be portrayed more negatively in Baidu Baike. We expect figures who are denounced by the CCP to be portrayed more negatively in Baidu Baike. On the other hand, we expect categories that are targets of positive propaganda to be portrayed more positively in Baidu Baike.  Overall, we expect that censorship and curation of Baidu Baike will mean that the words we are interested in will be treated similarly in Baidu Baike and state media outlet \emph{The People's Daily}. A summary of our theoretical expectations is presented in Table 1 below.

\begin{table}[htbp]
\caption{Theoretical Expections}
\begin{minipage}{\columnwidth}
\begin{center}
\begin{tabular}{@{}ll@{}}
\toprule
Category           & Sign             \\
\midrule
Freedom            & $-$   \\
Democracy          & $-$   \\
Election           & $-$  \\
Collective Action  & $-$   \\
Negative Figures   & $-$   \\
Social Control     & $+$   \\
Surveillance       & $+$   \\
CCP                & $+$  \\
Historical Events  & $+$   \\
Positive Figures   & $+$   \\
\bottomrule
\end{tabular}
\end{center}
\emph{Note:} Negative sign indicates Baidu Baike and \emph{People's Daily} are less favorable than Wikipedia and positive sign indicates that Baidu Baike and \emph{People's Daily} are more favorable than Wikipedia.
\end{minipage}
	
\end{table}
 
 \subsection{Limitations}
 
 Through this design, we test whether there are differences between word embeddings trained on Chinese language Wikipedia and those trained on Baidu Baike in topics where there is evidence of censorship on Baidu Baike.  While we think the evidence we produce is suggestive that censorship impacts the placement of the word embeddings, we cannot isolate the effect of censorship outside of other differences that may exist between Baidu Baike and Chinese language Wikipedia. Isolating the effect of censorship is difficult in part because censorship's influence is pervasive, affecting the content not only through pre-publication review, but also likely through the propensity for individuals to become editors and the information that they have and are willing to contribute.  This makes it very difficult to establish a counterfactual of what the content on Baidu Baike would have looked like without censorship.  We believe Chinese language Wikipedia is the closest approximation to this counterfactual.   

 \subsection{Results}
 
Following \citet{caliskan2017semantics}, we use a randomization test with one-sided p-value to compare how words in each category are represented in Wikipedia, Baidu Baike and \emph{People's Daily}.

Formally, let $X_{i}$, $i \in {a, b}$ be the set of word vectors for the target words from embedding $a$ and $b$ respectively. Let $A_{i}$, $B_{i}$, $i \in {a, b}$ be the two sets of word vectors for the attribute words, with $A$ being the set of positive attributes and $B$ being the set of negative attributes. Subscript $i$ again denotes the embedding that the word vectors are from. Let $\cos(\vec{p},\vec{q})$ denote the cosine of the angle between vectors $\vec{p}$ and $\vec{q}$. The test statistic is $$s_{i}(X, A, B) = \sum_{i \in a}s(x_{i}, A_{i}, B_{i}) - \sum_{i \in b}s(x_{i}, A_{i}, B_{i})$$
where
$$s(t, A, B) = \mbox{mean}_{p \in A}\cos(\vec{t},\vec{p}) - \mbox{mean}_{q \in B}\cos(\vec{t},\vec{q})$$

Let $\Omega$ denotes the set of all possible randomization realizations of assignment of word vector $x$ to embedding $i \in \{a, b\}$. The one-sided p-value of the permutation test is $$\mbox{Pr}_{i}[s_{\omega \in \Omega}(X, A, B) > s_{i}(X, A, B)]$$

We present the effect size of the difference in word associations across word embeddings, defined as 
$$\frac{\mbox{mean}_{i \in a}s(x_{i}, A_{i}, B_{i}) - \mbox{mean}_{i \in b}s(x_{i}, A_{i}, B_{i})}{\mbox{std.dev}_{i}s(x_{i}, A_{i}, B_{i})}$$ Conventional cutoffs for small, medium, and large effect sizes are $0.2$, $0.5$, and $0.8$, respectively. The comparisons between Wikipedia and Baidu Baike word embeddings and between Wikipedia and \emph{People's Daily} word embeddings are presented in Table 2 and Table 3 respectively.

\begin{table}
\centering
\caption{Wikipedia vs. Baidu Baike}
\resizebox{\columnwidth}{!}{\begin{tabular}{lrrrr}
\toprule
\multicolumn{1}{c}{ } & \multicolumn{2}{c}{Propaganda Attributes} & \multicolumn{2}{c}{Evaluative Attributes} \\
\cmidrule{2-3} \cmidrule{4-5}
  & effect size & p-value & effect size & p-value\\
\midrule
Freedom & -0.62 & 0.01 & 0.06 & 0.60\\
Democracy & -0.50 & 0.05 & -0.56 & 0.03\\
Election & -0.27 & 0.13 & -0.33 & 0.05\\
Collective Action & -0.66 & 0.00 & -0.09 & 0.34\\
Negative Figures & -0.91 & 0.00 & 0.50 & 0.99\\
\addlinespace
Social Control & 0.70 & 0.04 & 0.68 & 0.01\\
Surveillance & 0.09 & 0.32 & 0.73 & 0.00\\
CCP & 1.05 & 0.02 & 1.39 & 0.00\\
Historical Events & 0.14 & 0.19 & 0.27 & 0.01\\
Positive Figures & 0.59 & 0.00 & 1.17 & 0.00\\
\bottomrule
\end{tabular}}
\end{table}

\begin{table}
\centering
\caption{\label{tab:}Wikipedia vs. \emph{People's Daily}}
\resizebox{\columnwidth}{!}{\begin{tabular}{lrrrr}
\toprule
\multicolumn{1}{c}{ } & \multicolumn{2}{c}{Propaganda Attributes} & \multicolumn{2}{c}{Evaluative Attributes} \\
\cmidrule(l{3pt}r{3pt}){2-3} \cmidrule(l{3pt}r{3pt}){4-5}
  & effect size & p-value & effect size & p-value\\
\midrule
Freedom & -0.29 & 0.11 & -0.51 & 0.01\\
Democracy & -0.40 & 0.09 & -0.97 & 0.00\\
Election & -0.43 & 0.04 & -0.91 & 0.00\\
Collective Action & -0.81 & 0.00 & -0.10 & 0.34\\
Negative Figures & 0.44 & 0.91 & -0.06 & 0.41\\
\addlinespace
Social Control & 0.82 & 0.01 & 0.58 & 0.03\\
Surveillance & 0.31 & 0.06 & 0.84 & 0.00\\
CCP & 1.39 & 0.00 & 1.22 & 0.00\\
Historical Events & 0.29 & 0.08 & 0.22 & 0.04\\
Positive Figures & 1.51 & 0.00 & 1.29 & 0.00\\
\bottomrule
\end{tabular}}
\end{table}

Across most categories and for both sets of attribute words, the differences in word embeddings are in line with our theoretical expectations. Table 2 indicates that for categories Freedom, Democracy, Election, Collective Action, and Negative Figures, word embeddings trained with Baidu Baike display a more negative connotation than embeddings trained with Wikipedia. For categories Social Control, Surveillance, CCP, and Historical Events, word embeddings trained with Baidu Baike display a more positive connotation than embeddings trained with Wikipedia. The effect sizes indicate substantial differences for target words that are related to democracy and those that are targets of propaganda. This is consistent across both set of attribute words and across the two comparisons.  In Table 3 we show that the effect sizes when comparing Wikipedia and Baidu Baike are similar to comparing Wikipedia with the government publication \emph{The People's Daily}.

While most categories accord with our expectations, one in particular deserves further explanation.  Negative figures, including activists and dissidents who the CCP denounces, are only more significantly associated with negative words on Baidu Baike and \emph{People's Daily} in one instance and even have a positive effect size comparing Baidu Baike to Wikipedia in Table 2.  It is likely that because of censorship there is very little information about these figures in the Baidu Baike and \emph{People's Daily} corpuses, so their word embeddings do not show strong relationships with the attribute words. To examine this, we used Google Search to count the number of pages on Chinese language Wikipedia and Baidu Baike that link to each negative figure.  Out of 18 negative figures, Chinese language Wikipedia has more page links to two thirds of them, even though Chinese language Wikipedia is 16 times smaller.  Therefore, the uncertainty around the result we have for negative figures may be a result of lack of information about these individuals in Baidu Baike. 

%Additionally, target words related to freedom and election in a few of the tests are not statistically different across corpuses. Likely because China also declares it to be a democratic country with free and fair elections and some of these democratic concepts may be hijacked by the CCP to bolster its legitimacy.

%Also for historical events, just by virtue of many of the words in this category being associated with war, propaganda on these events may portray them as more brutal, causing more casualties, thus inadvertently creating negative associations for these word vectors. For example, for events such as “抗日战争”, propagandist narrative can portray the war as hard-fought and the CCP as suffering great losses but eventually overcoming the enemies in a brutal battle. Such a narrative, when used to train word embeddings, is likely to create some negative associations between the target word and the attributes. 

\section{Application: Sentiment Analysis of News Headlines}

In this section, we demonstrate that the differences we detected in word embeddings have tangible effect on downstream machine learning tasks.  To do this, we make use of the pre-trained word embeddings on each of the different corpuses as inputs in a larger machine learning model that automatically labels the sentiment polarity of news headlines. We chose the automated classification of news headlines because machine learning based on news headlines is used in recommendation systems for social media news feeds and news aggregators, as well as for analysts using automated classification of news to make stock price and economic predictions.\footnote{For example, EquBot \url{https://equbot.com/}.} We show that using the pre-trained word embeddings from Baidu Baike and Chinese language Wikipedia with identical training data produces sentiment predictions for news headlines that differ systematically across our categories of interest.

%Transfer learning using pre-trained word embeddings has enabled numerous breakthroughs in machine learning and artificial intelligence tasks (citation). In areas such as machine translation, speech recognition, predictive typing, and image recognition, transfer learning using pre-trained models has become the dominant approach. Specific to Word2Vec, it has been used for sentiment analysis, named entity recognition, part of speech tagging, among many other NLP tasks.

%Of the various NLP tasks, sentiment analysis of news has become widely used in automated news recommendation systems and for stock price predictions. In fact, sentiment analysis is a core component of many well-known companies such as Amazon and Google.

%To illustrate that the effect of censorship on Word2Vec models we documented in the previous sections have a tangible impact on real-life NLP applications, we show that using different Word2Vec models for the task of sentiment analysis generates sentiment labels of news headlines that differ systematically across categories of our interest.

\subsection{Data and Method}

We imagine a scenario where the task is to label the sentiment of news headlines where the model is trained on a large, general sample of news headlines.  We then examine the performance of this model on an oversample of headlines that include our target words.  This allows us to evaluate how a general news sentiment classifier performs on words that are politically valanced in China, varying the origin of the pre-trained embeddings, but holding constant the sentiment labels in the training and test sets.  

For the training set, we randomly select 5,000 headlines from the TNEWS dataset. The TNEWS dataset contains 73,360 Chinese news headlines of various categories.\footnote{For more details about the TNEWS dataset, see Appendix.} It is part of the Chinese Language Understanding Evaluation (CLUE) Benchmark and is widely used as the training data for Chinese news classification models. For each of the randomly selected 5,000 headlines, we label each news headline as positive, negative, or neutral in line with the general sentiment of the headline.  For our training set from the TNEWS dataset, we have 1,861 headlines with positive sentiment, 781 with negative sentiment, and 2,342 with neutral sentiment.\footnote{16 duplicated news headlines are dropped, resulting in 4,984 headlines in total.}

For the test set, we collect Chinese news headlines that contain any of our target words from Google News. For each of the target words, we collect up to 100 news headlines. Because some target words yield only a handful of news headlines, we collected 12,669 news headlines in total, out of 182 target words. Data collection was done in July and August of 2020. Using the exact same coding scheme as the training set, we label these headlines as positive, negative, or neutral.  The test set contains 5,291 headlines with positive sentiment, 3,913 with negative sentiment, and 3,424 with neutral sentiment.\footnote{41 duplicated news headlines are dropped, resulting in 12,628 headlines in total.}

We preprocess the news headlines by removing punctuation, numbers, special characters, the names of the news agency (if they appear on the headline), and duplicated headlines. To convert the news headlines into input for machine learning models, we first use a Chinese word segmentation tool to segment each news headline into a sequence of words. We then look up the word embedding for each word in the sequence. Following a conventional approach, we take the average of the pre-trained word embeddings of the words in a given news headline to represent each headline. Any word that does not have a corresponding word embedding in the Word2Vec models is dropped. This leaves us with three different representations of the headlines: one for Baidu Baike, one for Chinese language Wikipedia, and one for the \emph{People's Daily}.

With each of these three different representations of the text based on different pre-trained embeddings, we train three machine learning models -- Naive Bayes (NB), support vector machines (SVM) and TextCNN \citep{kim2014convolutional}.  For each model, we use identical training labels, from the TNEWS dataset.\footnote{Because headlines with neutral labels are more noisy and given the difficulty of training a three-class classifier with limited training data, we report results in the main text based on models that are trained with only positive and negative headlines. We report results with neutral headlines included in the Appendix. Our substantive conclusions are largely intact.}  This yields a total of nine models, with three for each pre-trained word embeddings. Each trained model is then used to predict sentiment labels on the test set. Because of the stochastic nature of TextCNN, the TextCNN results are averaged over 10 runs for each model.

We compare different trained models of the same architecture (NB, SVM, or TextCNN) by looking at the mis-classifications for each category of target words. Intuitively, a model that is pre-disposed to associate more positive words with a certain category of headlines will have more false-positives (e.g. negative headlines mis-classified as positive), whereas a model that is pre-disposed to associate more negative words with a certain category of headlines will have more false-negatives (e.g. positive headlines mis-classified as negative).

Because the overall mis-classification rate may differ for headlines of different target words, we use a linear mixed effects model to compare the different embeddings, allowing headlines of different target words to have different intercepts. More formally, let $L_{ij}$ be a list of $N$ human-labeled sentiment scores for headlines containing target word $i$ in category $j$. Let  $\hat{L}_{ij}^{a}$ and $\hat{L}_{ij}^{b}$ be the predicted sentiment scores from model $a$ and $b$ for the same headlines. We estimate the linear mixed effects model for each category $j$ of news headlines by
\begin{eqnarray}
y_{j} = \alpha_{ij} + X_{j}\beta_{j} + \epsilon_{j}
\label{mixedeffects}
\end{eqnarray}
where the outcome variable $y_{j}$ is a $2N \times 1$ vector of difference in classifications against human labels, 
$\big(\begin{smallmatrix}
  \hat{L}_{j}^{a} - L_{j}\\
  \hat{L}_{j}^{b} - L_{j}
\end{smallmatrix}\big)$.
$\alpha_{ij}$ is a $2N \times 1$ vector of random intercepts corresponding to headlines of each target word $i$ in category $j$. $X_{j}$ is an indicator variable for model $a$ (as opposed to $b$) and $\beta_{j}$ is the coefficient of interest.

\subsection{Results}

%\begin{enumerate}
%    \item The estimates are percentage point difference. An estimate of $-0.12$ means that if we compare the averages of the mis-classification rate $$\frac{\mbox{no. of false positives} - \mbox{no. of false negatives}}{\mbox{no. of headlines for category \textit{j}}}$$
    %the average mis-classification rate of the Baidu Baike (People's Daily) model is 12 percentage point lower than the Wikipedia model.
    %\item There are a number of other baseline models I have tried, like random forest, KNN, decision trees, and rbf svm. Results are generally similar. I simply picked the two models that are 1) commonly used and 2) have good accuracy in this case.
%\end{enumerate}

Before turning to the results of the impact of pre-trained embeddings on the predicted classifications of the model, we report the overall accuracy of each of the models on the test set in Table \ref{modelaccuracy}.  Overall, TextCNN performs the best out of the three models.  However, within models no set of pre-trained word embeddings performs better than the other -- they all perform quite similarly. 

\begin{table}[H]
\centering
\caption{Model Accuracy in Test Set \label{modelaccuracy}}
\begin{tabular}{llll}
\toprule
& Model & Accuracy\\
\midrule
{\textbf{Naive Bayes}} & &\\
& Baidu Baike & \hspace{1em}76.83\\
& Wikipedia & \hspace{1em}76.29\\
\midrule
{\textbf{SVM}} & &\\
& Baidu Baike & \hspace{1em}77.12\\
& Wikipedia & \hspace{1em}76.68\\
\midrule
{\textbf{TextCNN}} & &\\
& Baidu Baike & \hspace{1em}82.84\\
& Wikipedia & \hspace{1em}81.60\\
\bottomrule
\end{tabular}
\end{table}

Even though the selection of pre-trained embeddings does not seem to impact overall accuracy, the pre-trained embeddings do influence the false positive and false negative rates of different categories of headlines.  In Table \ref{wikibaiduclassifier} we show the comparison of Baidu Baike and Wikipedia, where Baidu Baike is model $a$ and Wikipedia is model $b$.  This means $X_j$ from Equation \ref{mixedeffects} is 1 for category $j$ if the model were trained with Baidu Baike word embeddings and 0 for Wikipedia.  A negative coefficient indicates that on average Baidu Baike rates this category more negatively than Wikipedia.  A positive coefficient indicates that on average Baidu Baike rates this category as more positive than Wikipedia.

\begin{table}[H]
\centering
\caption{Baidu Baike vs. Wikipedia\label{wikibaiduclassifier}}
\resizebox{\columnwidth}{!}{\begin{tabular}{lrrrrrr}
\toprule
\multicolumn{1}{c}{ } & \multicolumn{2}{c}{Naive Bayes} & \multicolumn{2}{c}{SVM} & \multicolumn{2}{c}{TextCNN} \\
\cmidrule(l{3pt}r{3pt}){2-3} \cmidrule(l{3pt}r{3pt}){4-5} \cmidrule(l{3pt}r{3pt}){6-7}
  & estimate & p-value & estimate & p-value & estimate & p-value\\
\midrule
Freedom & -0.13 & 0.00 & -0.06 & 0.00 & -0.04 & 0.04\\
Democracy & -0.08 & 0.00 & -0.05 & 0.04 & -0.04 & 0.06\\
Election & -0.11 & 0.00 & -0.06 & 0.03 & -0.02 & 0.48\\
Collective Action & -0.13 & 0.00 & -0.07 & 0.00 & -0.05 & 0.01\\
Negative Figures & -0.04 & 0.03 & 0.00 & 0.96 & -0.01 & 0.54\\
\addlinespace
Social Control & 0.03 & 0.12 & 0.00 & 0.93 & 0.03 & 0.13\\
Surveillance & -0.01 & 0.68 & -0.01 & 0.80 & 0.00 & 0.91\\
CCP & 0.03 & 0.21 & 0.01 & 0.65 & 0.03 & 0.05\\
Historical Events & -0.04 & 0.04 & 0.01 & 0.75 & -0.02 & 0.26\\
Positive Figures & 0.06 & 0.00 & 0.06 & 0.00 & 0.06 & 0.00\\
\bottomrule
\end{tabular}}
\end{table}

The results are largely consistent with what we found in Section \ref{sec:distance}. Overwhelmingly, Wikipedia predicts headlines that contain target words in the categories of freedom, democracy, election, and collective action to be more positive. In contrast, Baidu Baike predicts headlines that contain target words of figures that the CCP views positively to be more positive. The exceptions to our expectations are the categories of social control, surveillance, CCP, and historical events, where we cannot reject the null of no difference between the two corpuses, although they do not go against our expectations.  We find similar results for the comparison between \emph{People's Daily} and Chinese language Wikipedia, in Table \ref{peoplesdailywikiclassifier}.  

\begin{table}[H]
\centering
\caption{\emph{People's Daily} vs. Wikipedia \label{peoplesdailywikiclassifier}}
\resizebox{\columnwidth}{!}{\begin{tabular}{lrrrrrr}
\toprule
\multicolumn{1}{c}{ } & \multicolumn{2}{c}{Naive Bayes} & \multicolumn{2}{c}{SVM} & \multicolumn{2}{c}{TextCNN} \\
\cmidrule(l{3pt}r{3pt}){2-3} \cmidrule(l{3pt}r{3pt}){4-5} \cmidrule(l{3pt}r{3pt}){6-7}
  & estimate & p-value & estimate & p-value & estimate & p-value\\
\midrule
Freedom & -0.22 & 0.00 & -0.08 & 0.00 & -0.12 & 0.00\\
Democracy & -0.14 & 0.00 & -0.06 & 0.02 & -0.07 & 0.00\\
Election & -0.13 & 0.00 & -0.01 & 0.62 & -0.04 & 0.12\\
Collective Action & -0.19 & 0.00 & -0.05 & 0.05 & -0.06 & 0.00\\
Negative Figures & 0.01 & 0.78 & 0.01 & 0.72 & -0.05 & 0.01\\
\addlinespace
Social Control & 0.05 & 0.00 & 0.01 & 0.66 & 0.01 & 0.63\\
Surveillance & -0.04 & 0.11 & -0.02 & 0.34 & -0.03 & 0.22\\
CCP & 0.07 & 0.00 & 0.00 & 0.82 & 0.02 & 0.24\\
Historical Events & -0.01 & 0.77 & 0.02 & 0.29 & -0.01 & 0.44\\
Positive Figures & 0.13 & 0.00 & 0.04 & 0.00 & 0.06 & 0.00\\
\bottomrule
\end{tabular}}
\end{table}

To provide intuition, Figure \ref{fig:example} shows examples of headlines labeled differently between model trained with Baidu Baike pre-trained embeddings and model trained with Chinese language Wikipedia in our test set. The model trained with Baidu Baike pre-trained word embedding labeled ``Tsai Ing-wen: Hope Hong Kong Can Enjoy Democracy as Taiwan Does" as negative, while Wikipedia and humans labeled this headline as positive.  The difference in these predictions do not stem from the training data -- which is the same -- or the model -- which is the same.  Instead, the associations made within the pre-trained word embeddings drive these differences.

\begin{CJK*}{UTF8}{gbsn}

\begin{figure}[htbp]
\fbox{
\begin{minipage}{\columnwidth}
Example 1: 蔡英文: 盼台湾享有的民主自由香港也可以有\\
Tsai Ing-wen: Hope Hong Kong Can Enjoy Democracy as Taiwan Does\\
Baidu Baike Label: - \quad Wikipedia Label: + \quad Human Label: + \\
\hrule
Example 2: 封杀文化席卷欧美\mbox{ }自由反被自由误?\\
Cancel Culture Spreading through the Western World, Is It the Fault of Freedom?\\
Baidu Baike Label: - \quad Wikipedia Label: + \quad Human Label: - \\
\hrule
Example 3: 共产暴政录: 抗美援朝真相\\
Communist Tyranny: The Truth about Chinese Involvement in\\
the Korean War\\
Baidu Baike Label: + \quad Wikipedia Label: - \quad Human Label: - \\
\hrule
Example 4: 香港《国安法》：中国驻港部队司令强硬表态维稳\\
Hong Kong Security Law: PLA Hong Kong Garrison Commander Takes Tough
Stance in Support of Stability Maintenance\\
Baidu Baike Label: + \quad Wikipedia Label: - \quad Human Label: - \\
\end{minipage}
}
\Description{Examples of Headlines Labeled Differently By Naive Bayes Models Trained with Baidu Baike and Wikipedia}
\caption{Examples of Headlines Labeled Differently By Naive Bayes Models Trained with Baidu Baike and Wikipedia}
\label{fig:example}
\end{figure}
\end{CJK*}

\section{Conclusion}                      
The extensive use of censorship in China means that the Chinese government is in the dominant position to shape the political content of large Chinese language corpuses. Even though corpuses like Chinese language Wikipedia exist outside of the Great Firewall, they are significantly weakened by censorship, as shown by the smaller size of Chinese language Wikipedia in comparison to Baidu Baike. While more work would need to be done to understand how these discrepancies affects users of any particular application, we showed in this paper that political differences reflective of censorship exist between two of the corpuses commonly used to train Chinese language NLP. While our work focuses on word embeddings, the discrepancies we uncovered likely affect other pre-trained NLP models as well, such as BERT \citep{devlin2018bert} and ERNIE \citep{sun2019ernie}. Furthermore, these political differences present a pathway through which political censorship can have downstream effects on applications that may not themselves be political but that rely on NLP, from predictive text and article recommendation systems to social media news feeds and algorithms that flag disinformation.

The literature in computer science has taken on the problem of bias in training data by looking for ways to de-bias it -- for example, through data augmentation \citep{zhao2018gender}, de-biasing word embeddings \citep{bolukbasi2016man}, and adversarial learning \citep{zhang2018mitigating}.\footnote{Although methods for de-biasing have also been shown to often be inadequate \citep{gonen2019lipstick, blodgett2020language}.} However, it is unclear how to think about de-biasing attitudes toward democracy, freedom, surveillance, and social control.  What does unbiased look like in these circumstances, and how would one test it?  The only way we can think about an unbiased training set in this circumstance is one where certain ideas are not automatically precluded from being included in any given corpus. But knowing what perspectives have been omitted is difficult to determine and correct after the fact.

%More research should be done on the way in which governments can shape the data that are used to in AI applications and the impact of these actions on user welfare and behavior.   %Among the challenges of studying this is knowing the underpinnings of algorithms we use in data to day life and being unable to experiment on them \citep{rahwan2019machine}.  

\begin{acks}
This work is partially supported by the \grantsponsor{RIDIR000}{National Science Foundation}{https://www.nsf.gov/awardsearch/showAward?AWD_ID=1738411} under Grant
No.:˜\grantnum{RIDIR}000{1738411}.  Thanks to Guanwei Hu, Yucong Li, and Zoey Jialu Xu for their excellent research assistance.  We thank Michelle Torres, Allan Dafoe, and Jeffrey Ding for their helpful comments on this work.
\end{acks}


\bibliographystyle{ACM-Reference-Format}
\bibliography{fff}

\appendix
\counterwithin{figure}{section}
\counterwithin{table}{section}

\pagebreak
%\section{Examples of censorship in Baidu Baike}
%\begin{figure}[htbp]%
	%\centering
	%\includegraphics[width=0.6\linewidth]{liusi.png}
	%\caption{Attempt to create an entry for the ``June Fourth (Movement)" was prohibited by Baidu Baike}
	%\label{fig:liusi}
%\end{figure}
%\begin{figure}[htbp]%
	%\centering
	%\includegraphics[width=0.6\linewidth]{wuerkaixi.png}
	%\caption{Attempt to create an entry for ``Wu'erkaixi" was prohibited by Baidu Baike}
	%\label{fig:wuerkaixi}
%\end{figure}

\section{Additional Sentiment Analysis Results}

\subsection{Model Accuracy on Validation Set}
In training the TextCNN models, we held out $20\%$ of our training set as a validation set. The validation set was used to assess the quality of the models during training. The model with the best accuracy on the validation set in each run was selected as the outputted model. \ref{appendix_modelaccuracyvalidation} reports the average accuracy (over 10 runs) of the models on the validation sets.

\begin{table}[H]
\caption{Model Accuracy on Validation Sets}
\label{appendix_modelaccuracyvalidation}
\begin{minipage}{\columnwidth}
\begin{center}
\begin{tabular}{llll}
\toprule
{\textbf{2-class}} & &\\
& Baidu Baike & \hspace{1em}90.29\\
& Wikipedia & \hspace{1em}89.65\\
& People's Daily & \hspace{1em}92.64\\
\midrule
{\textbf{3-class}} & &\\
& Baidu Baike & \hspace{1em}67.44\\
& Wikipedia & \hspace{1em}66.07\\
& People's Daily & \hspace{1em}67.80\\
\bottomrule
\end{tabular}
\end{center}
\emph{Note:} ``2-class" classification means that the training and validation sets contain only negative and positive headlines. ``3-class" classification additionally has neutral headlines included.
\end{minipage}
\end{table}

\subsection{Sentiment Analysis Results with Neutral Headlines Included}
\begin{table}[H]
\centering
\caption{Model Accuracy}
\begin{tabular}{llll}
\toprule
& Model & Accuracy\\
\midrule
{\textbf{Naive Bayes}} & &\\
& Baidu Baike & \hspace{1em}56.42\\
& Wikipedia & \hspace{1em}55.63\\
& People's Daily & \hspace{1em}57.79\\
\midrule
{\textbf{SVM}} & &\\
& Baidu Baike & \hspace{1em}55.53\\
& Wikipedia & \hspace{1em}55.29\\
& People's Daily & \hspace{1em}54.71\\
\midrule
{\textbf{TextCNN}} & &\\
& Baidu Baike & \hspace{1em}61.71\\
& Wikipedia & \hspace{1em}60.89\\
& People's Daily & \hspace{1em}58.55\\
\bottomrule
\end{tabular}
\end{table}

\begin{table}[H]
\centering
\caption{Wikipedia vs. Baidu Baike}
\resizebox{\columnwidth}{!}{\begin{tabular}{lrrrrrr}
\toprule
\multicolumn{1}{c}{ } & \multicolumn{2}{c}{Naive Bayes} & \multicolumn{2}{c}{SVM} & \multicolumn{2}{c}{TextCNN} \\
\cmidrule(l{3pt}r{3pt}){2-3} \cmidrule(l{3pt}r{3pt}){4-5} \cmidrule(l{3pt}r{3pt}){6-7}
  & estimate & p-value & estimate & p-value & estimate & p-value\\
\midrule
Freedom & -0.11 & 0.00 & -0.06 & 0.00 & -0.03 & 0.12\\
Democracy & -0.08 & 0.00 & -0.04 & 0.04 & -0.02 & 0.23\\
Election & -0.09 & 0.00 & 0.00 & 0.87 & -0.01 & 0.62\\
Collective Action & -0.10 & 0.00 & -0.06 & 0.00 & 0.00 & 0.89\\
Negative Figures & -0.05 & 0.00 & -0.01 & 0.47 & 0.03 & 0.02\\
\addlinespace
Social Control & 0.01 & 0.59 & 0.03 & 0.08 & 0.03 & 0.04\\
Surveillance & -0.06 & 0.00 & -0.05 & 0.00 & 0.01 & 0.51\\
CCP & 0.05 & 0.00 & 0.03 & 0.01 & 0.04 & 0.01\\
Historical Events & -0.04 & 0.02 & -0.01 & 0.66 & 0.02 & 0.05\\
Positive Figures & 0.08 & 0.00 & 0.07 & 0.00 & 0.08 & 0.00\\
\bottomrule
\end{tabular}}
\end{table}

\begin{table}[H]
\centering
\caption{Wikipedia vs. People's Daily}
\resizebox{\columnwidth}{!}{\begin{tabular}{lrrrrrr}
\toprule
\multicolumn{1}{c}{ } & \multicolumn{2}{c}{Naive Bayes} & \multicolumn{2}{c}{SVM} & \multicolumn{2}{c}{TextCNN} \\
\cmidrule(l{3pt}r{3pt}){2-3} \cmidrule(l{3pt}r{3pt}){4-5} \cmidrule(l{3pt}r{3pt}){6-7}
  & estimate & p-value & estimate & p-value & estimate & p-value\\
\midrule
Freedom & -0.17 & 0.00 & -0.07 & 0.00 & -0.05 & 0.01\\
Democracy & -0.13 & 0.00 & -0.07 & 0.00 & -0.06 & 0.00\\
Election & -0.13 & 0.00 & 0.00 & 0.93 & -0.01 & 0.53\\
Collective Action & -0.15 & 0.00 & -0.06 & 0.00 & -0.02 & 0.22\\
Negative Figures & -0.02 & 0.17 & 0.00 & 0.96 & 0.01 & 0.32\\
\addlinespace
Social Control & 0.05 & 0.00 & 0.02 & 0.22 & 0.00 & 0.97\\
Surveillance & -0.01 & 0.61 & -0.04 & 0.02 & -0.01 & 0.56\\
CCP & 0.04 & 0.01 & 0.04 & 0.00 & 0.03 & 0.02\\
Historical Events & -0.01 & 0.53 & 0.00 & 0.78 & 0.03 & 0.00\\
Positive Figures & 0.10 & 0.00 & 0.06 & 0.00 & 0.10 & 0.00\\
\bottomrule
\end{tabular}}
\end{table}

\subsection{Sentiment Analysis Results Comparing Baidu Baike and People's Daily}
\ref{appendix_baidupd2classclassifier} reports the results comparing models trained on Baidu Baike and those trained on People's Daily, where Baidu Baike is model $a$ and People's Daily is model $b$. A positive coefficient means that on average People's Daily model rates a given category more positively than Baidu Baike.

\ref{appendix_baidupd3classclassifier} reports results from the same comparison but with headlines with neutral labels included in the training and test sets. 

\begin{table}[H]
\centering
\caption{Baidu Baike vs. People's Daily (2-class)}
\resizebox{\columnwidth}{!}{\begin{tabular}{lrrrrrr}
\toprule
\multicolumn{1}{c}{ } & \multicolumn{2}{c}{Naive Bayes} & \multicolumn{2}{c}{SVM} & \multicolumn{2}{c}{TextCNN} \\
\cmidrule(l{3pt}r{3pt}){2-3} \cmidrule(l{3pt}r{3pt}){4-5} \cmidrule(l{3pt}r{3pt}){6-7}
  & estimate & p-value & estimate & p-value & estimate & p-value\\
\midrule
Freedom & -0.09 & 0.00 & -0.02 & 0.48 & -0.07 & 0.00\\
Democracy & -0.05 & 0.05 & -0.01 & 0.68 & -0.02 & 0.29\\
Election & -0.03 & 0.31 & 0.04 & 0.08 & -0.02 & 0.36\\
Collective Action & -0.06 & 0.01 & 0.02 & 0.28 & -0.01 & 0.57\\
Negative Figures & 0.05 & 0.02 & 0.01 & 0.69 & -0.04 & 0.04\\
\addlinespace
Social Control & 0.03 & 0.09 & 0.01 & 0.72 & -0.02 & 0.27\\
Surveillance & -0.03 & 0.25 & -0.02 & 0.49 & -0.02 & 0.24\\
CCP & 0.04 & 0.04 & 0.00 & 0.82 & -0.01 & 0.33\\
Historical Events & 0.04 & 0.07 & 0.01 & 0.46 & 0.01 & 0.72\\
Positive Figures & 0.07 & 0.00 & -0.01 & 0.35 & 0.00 & 0.92\\
\bottomrule
\end{tabular}}
\label{appendix_baidupd2classclassifier}
\end{table}

\begin{table}[H]
\centering
\caption{Baidu Baike vs. People's Daily (3-class)}
\resizebox{\columnwidth}{!}{\begin{tabular}{lrrrrrr}
\toprule
\multicolumn{1}{c}{ } & \multicolumn{2}{c}{Naive Bayes} & \multicolumn{2}{c}{SVM} & \multicolumn{2}{c}{TextCNN} \\
\cmidrule(l{3pt}r{3pt}){2-3} \cmidrule(l{3pt}r{3pt}){4-5} \cmidrule(l{3pt}r{3pt}){6-7}
  & estimate & p-value & estimate & p-value & estimate & p-value\\
\midrule
Freedom & -0.07 & 0.00 & -0.01 & 0.64 & -0.02 & 0.21\\
Democracy & -0.06 & 0.01 & -0.03 & 0.17 & -0.04 & 0.04\\
Election & -0.04 & 0.07 & 0.00 & 0.93 & 0.00 & 0.88\\
Collective Action & -0.06 & 0.00 & 0.00 & 0.84 & -0.02 & 0.26\\
Negative Figures & 0.03 & 0.07 & 0.01 & 0.44 & -0.02 & 0.20\\
\addlinespace
Social Control & 0.04 & 0.02 & -0.01 & 0.59 & -0.03 & 0.04\\
Surveillance & 0.05 & 0.00 & 0.01 & 0.55 & -0.02 & 0.20\\
CCP & -0.01 & 0.63 & 0.01 & 0.46 & 0.00 & 0.73\\
Historical Events & 0.03 & 0.06 & 0.00 & 0.88 & 0.01 & 0.36\\
Positive Figures & 0.02 & 0.01 & -0.01 & 0.34 & 0.01 & 0.12\\
\bottomrule
\end{tabular}}
\label{appendix_baidupd3classclassifier}
\end{table}

\section{Further Details on the TNEWS Dataset}
The TNEWS Dataset comprises of 73,360 Chinese news headlines from Toutiao, a Chinese news and information content platform. The dataset contains news headlines from 15 categories: \textit{story, culture, entertainment, sports, finance, house, car, education, technology, military, travel, world, stock, agriculture} and \textit{gaming}.

The TNEWS dataset is part of the Chinese Language Understanding Evaluation (CLUE) Benchmark, which serves as a common repository of datasets used to test the accuracy of trained models. (For an equivalent of CLUE in English, see GLUE: \url{https://gluebenchmark.com/}). Because the length of a news headline is usually short, the TNEWS dataset is widely used as either training or testing data for machine learning models that tackle short-text classification tasks. Given that the downstream task we are interested in is the classification of news headlines, the TNEWS dataset serves as the ideal source of data in our case.

The TNEWS dataset is split into a training set (53,360 headlines), a validation set (10,000 headlines) and a test set (10,100 headlines). For our purpose, we pooled the three sets and randomly selected 5,000 news headlines from the pooled set. Because the news headlines are not labeled according to sentiment in the dataset, we manually labeled the sentiment of the headlines in our selected subset. Each headline is labeled by two independent coders of native Chinese speaker and any conflict in labeling is resolved.

\section{List of Target Words}
\begin{CJK*}{UTF8}{gbsn}
Freedom (自由) = \{自由 (freedom), 言论自由 (freedom of speech), 集会自由 (freedom of assembly), 新闻自由 (freedom of the press), 结社自由 (freedom of association), 自由权 (right to freedom), 民主自由 (democracy and freedom), 自由言论 (free speech), 创作自由 (creative freedom), 婚姻自主 (marital autonomy), 自由民主 (freedom and democracy), 自由市场 (free market), 自决 (self-determination), 自决权 (right to self-determination), 生而自由 (born free), 自由自在 (free), 自由选择 (freedom of choice), 自由思想 (freedom of thought), 公民自由 (civil liberties), 自由竞争 (free competition), 宗教自由 (freedom of religion), 自由价格 (free price)\}\\

\noindent Election (选举) = \{选举 (election), 直接选举 (direct election), 议会选举 (parliamentary election), 间接选举 (indirect election), 直选 (direct election), 换届选举 (general election), 民选 (democratically elected), 投票选举 (voting), 全民公决 (referendum), 总统大选 (presidential election), 大选 (election), 普选 (universal suffrage), 全民投票 (referendum), 民主选举 (democratic election)\}\\

\noindent Democracy (民主) = \{民主 (democracy), 自由民主 (freedom and democracy), 民主自由 (democracy and freedom), 民主制度 (democratic system), 民主化 (democratization), 社会民主主义 (social democracy), 民主运动 (democratic movement), 民主主义 (democracy) , 民主改革 (democratic reform), 民主制 (democratic system), 民主选举 (demoratic election), 民主权力 (democratic rights), 多党制 (multi-party system), 民主法制 (democracy and rule of law), 民主权利 (democratic rights)\}\\

\noindent Social Control (维稳) = \{维稳 (social control), 处突 (emergency handling), 社会治安 (public security), 反恐怖 (counter-terrorism), 公安工作 (police work), 预防犯罪 (crime prevention), 收容审查 (arrest and investigation), 治安工作 (public security work), 大排查 (inspections), 扫黄打非 (combating pornography and illegal publications), 接访 (petition reception), 反邪教 (anti-cult)\}\\

\noindent Surveillance (监控) = \{监控 (surveillance), 监测 (monitor), 监视 (surveillance), 管控 (control), 监看 (monitor), 监视系统 (surveillance system), 截听 (tapping), 监控中心 (surveillance center), 情报服务 (intelligence service), 排查 (inspection), 监视器 (surveillance equipment), 情报搜集 (intelligence collection), 间谍卫星 (reconnaissance satellite) , 管理网络 (internet control), 监控器 (surveillance equipment), 监控站 (surveillance center), 监控室 (surveillance center), 数据采集 (data collection)\}\\

\noindent Collective Action (抗议) = \{抗议 (protest), 示威 (demonstration), 示威游行 (demonstration; march), 示威抗议  ( demonstration; protest), 游行示威 (demonstration; march), 静坐示威 (sit-in), 绝食抗议 (hunger strike), 请愿 (petition), 示威运动 (demonstration), 游行 (demonstration; march), 罢教 (strike), 静坐 (sit-in), 集会游行 (demonstration; assembly), 罢课 (strike), 签名运动 (signature campaign)\}\\

\noindent Positive Figures (党和国家) = \{毛泽东 (Mao Zedong), 江泽民 (Jiang Zemin), 胡锦涛 (Ju Jintao), 习近平 (Xi Jinping), 周恩来 (Zhou Enlai), 朱镕基 (Zhu Rongji), 温家宝 (Wen Jiabao), 李克强 (Li Keqiang), 邓小平 (Deng Xiaoping), 曾庆红 (Zeng Qinghong), 华国锋 (Hua Guofeng), 李鹏 (Li Peng), 杨尚昆 (Yang Shangkun), 谷牧 (Gu Mu), 吴邦国 (Wu Bangguo), 李岚清 (Li Lanqing), 纪登奎 (Ji Dengkui), 乔石 (Qiao Shi), 邹家华 (Zou Jiahua), 李瑞环 (Li Ruihuan), 俞正声 (Yu Zhengsheng), 张高丽 (Zhang Haoli), 田纪云 (Tian Jiyun), 回良玉 (Hui Liangyu), 李源潮 (Li Yuanchao), 贾庆林 (Jia Qinglin), 姚依林 (Yao Yilin), 张立昌 (Zhang Lichang), 尉健行 (Wei Jianxing), 姜春云 (Jiang Chunyun), 李铁映 (Li Tieying), 王兆国 (Wang Zhaoguo), 罗干 (Luo Gan), 刘靖基 (Liu Jingji), 杨汝岱 (Yang Rudai), 王光英 (Wang Guangying), 彭佩云 (Peng Peiyun), 刘云山 (Liu Yunshan), 丁关根 (Ding Guangen), 彭真 (Peng Zhen), 胡启立 (Hu Qili), 曾培炎 (Zeng Peiyan), 何东昌 (He Dongchang)\}\\

\noindent Negative Figures = \{林彪 (Lin Biao), 王洪文 (Wang Hongwen), 张春桥 (Zhang Chunqiao), 江青 (Jiang Qing), 姚文元 (Yao Wenyuan), 刘晓波 (Liu Xiaobo), 丹增嘉措 ( Tenzin Gyatso), 李洪志 ( Li Hongzhi), 陈水扁 (Chen Shui-bian), 黄之锋 (Joshua Wong), 黎智英 (Jimmy Lai), 艾未未 (Ai Weiwei), 李登辉 (Lee Teng-hui), 李柱铭 (Martin Lee), 何俊仁 (Albert Ho), 陈方安生 (Anson Chan), 达赖 (Dalai Lama), 陈光诚 (Chen Guangcheng), 滕彪 (Teng Biao), 魏京生 (Wei Jingsheng), 鲍彤 (Bao Tong)\}\\

\noindent CCP (中国共产党) = \{党中央 (central committee), 中国共产党 (CCP), 党支部 (party branch), 中共中央 (central committee), 共青团 (CCP youth league), 共青团中央 (youth league central committee), 党委 (party committee), 中央党校 (central party school)\}\\

\noindent Historical Events = \{抗日战争 (Anti-Japanese War), 解放战争 (China's War of Liberation), 抗美援朝 (the War to resist U.S. Aggression and Aid Korea), 改革开放 (Reform and Opening up), 香港回归 (Hong Kong reunification), 长征 (Long March), 三大战役 (Three Great Battles in the Second Civil War), 秋收起义 (Autumn Harvest Uprising), 南昌起义 (Nanchang Uprising), 澳门回归 (Transfer of sovereignty over Macau), 志愿军 (Volunteer Army), 土地改革 (Land Reform), 六四 (June Fourth Movement), 遵义会议 (Zunyi Conference), 九二南巡 (Deng's Southern Tour in 1992), 广州起义 (Guangzhou Uprising), 西藏和平解放 (Annexation of Tibet), 井冈山会师 (Jinggangshan Huishi), 百团大战 (Hundred Regiments Offensive), 文革 (Cultural Revolution), 文化大革命 (Cultural Revolution), 大跃进 (Great Leap Forward), 四人帮 (Gang of Four), 解放农奴 (Serfs Emancipation)\}
\end{CJK*}
\section{Lists of Propaganda Attribute Words}
\begin{CJK*}{UTF8}{gbsn}
\noindent Positive Adjectives = \{稳定, 繁荣, 富强, 平稳, 幸福, 振兴, 发展, 兴旺, 昌盛, 强盛, 稳当, 安定, 局势稳定, 安定团结, 长治久安, 安居乐业\}\\

\noindent Negative Adjectives = \{动荡, 衰落, 震荡, 贫瘠, 不幸,  衰退, 萧条, 败落, 没落, 衰败, 摇摆, 不稳, 时局动荡, 颠沛流离, 动荡不安, 民不聊生\}\\
\end{CJK*}
\section{Examples of Evaluative Attribute Words}
\begin{CJK*}{UTF8}{gbsn}
\noindent Positive Evaluative = \{情投意合, 精选, 严格遵守, 最根本, 确有必要, 重镇, 直接接管, 收获, 思想性, 均需参加, 可用于, 当你落后, 同意接受, 居冠, 感化, 完美演出, 急欲, 多元地理环境, 形影不离的朋友, 一举击败, \dots\}\\

\noindent Negative Evaluative = \{金融波动, 科以, 畸型, 向..开枪, 破碎家庭, 撬动, 头皮发麻, 颠覆, 迟疑, 血淋淋地, 驱赶, 干的好事, 责骂不休, 生硬, 沖蚀, 拉回, 走失的家畜, 燃眉之急, 喷溅, 违反, \dots\}\\

For the full list of evaluative words from the augmented NTU sentiment dictionary (ANTUSD), see \url{https://academiasinicanlplab.github.io/#resources}.
\end{CJK*}
\balance
\end{document}